\pdfoutput=1
\documentclass[11pt]{article}
\usepackage{authblk}
\usepackage[final]{acl}
\usepackage{times}
\usepackage{latexsym}
\usepackage{amsmath}
\usepackage[T1]{fontenc}
\usepackage{booktabs}
\usepackage{url}
\usepackage{float}
\usepackage[utf8]{inputenc}
\usepackage{microtype}
 \usepackage{graphicx}
\title{Generating multiple-choice questions for medical question answering\\
with distractors and cue-masking}
\newcommand{\blue}[1]{\textcolor{black}{#1}}

\author[1]{Damien Sileo}
\author[2]{Kanimozhi Uma}
\author[2]{Marie-Francine Moens}

\affil[1]{Univ. Lille, Inria, CNRS, Centrale Lille, UMR 9189 - CRIStAL, F-59000 Lille, France}

\affil[2]{Department of Computer Science, KU Leuven, Belgium} \vspace{-2ex}
\affil[ ]{\url{damien.sileo@inria.fr}}

\usepackage[subtle]{savetrees}

\newcommand\footnotehref[2]{\footnote{\href{#1}{\texttt{#2}}}}

\begin{document}
\maketitle
\begin{abstract}
Medical multiple-choice question answering (MCQA) is particularly difficult. Questions may describe patient symptoms and ask for the correct diagnosis, which  requires domain knowledge and complex reasoning. Standard language modeling pretraining alone is not sufficient to achieve the best results.  \citet{jin2020disease} showed that focusing masked language modeling on disease name prediction when using medical encyclopedic paragraphs as input leads to considerable MCQA accuracy improvement. In this work, we show that (1) fine-tuning on generated MCQA dataset outperforms the masked language modeling based objective and (2) correctly masking the cues to the answers is critical for good performance. We release new pretraining datasets and achieve state-of-the-art results on 4 MCQA datasets, notably  +5.7\% with base-size model on MedQA-USMLE.

\end{abstract}

\section{Introduction}

The multiple-choice question answering \blue{\cite{rogers2021qa}} task can be formulated with $\{ \text{Q}, \{ \text{O}_1 ...\text{O}_N\}\}$ examples  where Q represents a question and O the candidate options. The goal is to select the correct answer from the options. Medical multiple-choice question answering \blue{(MCQA)} has valuable applications for patient or physician assistance, but is limited by the accuracy of current systems. 
Medical knowledge is key to this task which can ask questions about patient diagnosis or the appropriate treatment. Medical knowledge graphs, such as UMLS \cite{umls93} and SnomedCT \cite{donnelly2006snomed} mainly encode terminological knowledge \citep{SCHULZ2001207}. \blue{They} are sparse when it comes to the practical medical knowledge which is instead available as text in encyclopedias. Various training methodologies allow text encoders to absorb external knowledge.  Text encoders acquire some factual knowledge via masked language modeling (MLM) \citep{petroni-etal-2019-language}, but \citet{jin2020disease} showed that MLM objective \blue{solely focused} on disease names significantly enhances downstream task accuracy.

We compare targeted MLM with auxiliary pretraining on a generated MCQA dataset constructed with medical concepts as answers, associated paragraphs as questions Q, and generated distractors as other options. In particular, we show that we can leverage \textit{differential diagnoses} to obtain distractors. Strictly speaking,  differential diagnosis is the process of differentiating several conditions by examining the associated clinical features with additional tests. The term differential diagnosis is also used to denote commonly associated conditions that often need to be distinguished – \textit{bronchitis} is a differential diagnosis of \textit{common cold}.
We assemble a dataset of differential diagnoses and show that they provide helpful distractors. We also show that we can find differential diagnoses with a 
model trained to retrieve incorrect options based on the correct option from an existing MCQA dataset.

We then analyze the importance of properly masking the cues\footnote{\blue{A cue is the presence of a set of tokens that can help the prediction of the correct answer.}} to the correct option O$^*$, i.e., parts of the answer that are present in Q. The DiseaseBERT pretraining masks all the tokens from the disease name to incentivize the model to look at the symptoms. We show that token-level masking is sub-optimal, as the masking of some tokens can also give away the answer. We propose a new masking scheme tailored to MCQA pretraining, called probability-matching cue masking, to prevent both present and masked token from giving away the answer. We also collect new sources of encyclopedic text for medical pretraining. Our contributions are: (i) We compare MLM, targeted MLM and auxiliary fine-tuning on generated MCQA data;  (ii) We identify issues with previous cue-masking techniques and propose a new masking strategy; (iii) We propose and distribute\footnote{\href{https://huggingface.co/datasets/sileod/wikimedqa}{\texttt{hf.co/datasets/sileod/wikimedqa}}} new pretraining datasets; and (iv) We perform controlled comparison experiments for our contributions and achieve state-of-the-art on 4 datasets.

\section{Related work}
\paragraph{Medical text encoders pretraining} 
Numerous models
\cite{lewis-etal-2020-pretrained, lee2020biobert, michalopoulos-etal-2021-umlsbert, lewis-etal-2020-pretrained, kanakarajan-etal-2021-bioelectra, gu2021domain, yasunaga2022linkbert} adapt BERT pretraining to the biomedical domain to derive domain-specific text encoders.
Our work is close to DiseaseBERT \cite{he-etal-2020-infusing} which builds upon these encoders as an additional pretraining stage \cite{phang2018sentence} to improve their knowledge. 
\blue{Other work focus on external knowledge extraction \cite{rwextractive, rwknowledge} and integration \citep{Realm,kadapter}, but knowledge augmented models still rely on pretrained text encoders.}

\paragraph{Medical question answering} Multiple datasets were proposed for medical MCQA for English \cite{jin2020disease, pmlr-v174-pal22a,hendryckstest2021}, Spanish (with English translations) \citep{vilares-gomez-rodriguez-2019-head} and 
Chinese \cite{jin2020disease, li-etal-2021-mlec}. PubMedQA \cite{jin-etal-2019-pubmedqa} and emrQA \cite{pampari-etal-2018-emrqa} are other large-scale biomedical QA datasets, but they address extractive question answering, i.e., cases where the answer to a question is explicitly in the text. 
Our work is the first to generate MCQA data for medical domain pretraining.

\paragraph{Distractor prediction} Our work is related to the problem of generating distractors for multiple-choice questions.
These models use the existing answers to derive other answers that are plausible yet wrong. We distinguish two strands of approaches. Retrieval-based models use the correct answer as a query to retrieve related yet wrong alternatives among the answers to other questions \cite{ha-yaneva-2018-automatic}.
Generation-based models \cite{chung-etal-2020-BERT} learn to generate distractors with language models and focus on the  diversity and adequacy of the generated distractors. Here, we tailor distractor prediction to medical MCQA and also draw a new parallel between distractor generation and differential diagnosis.

\begin{figure}
\includegraphics[width =0.47\textwidth]{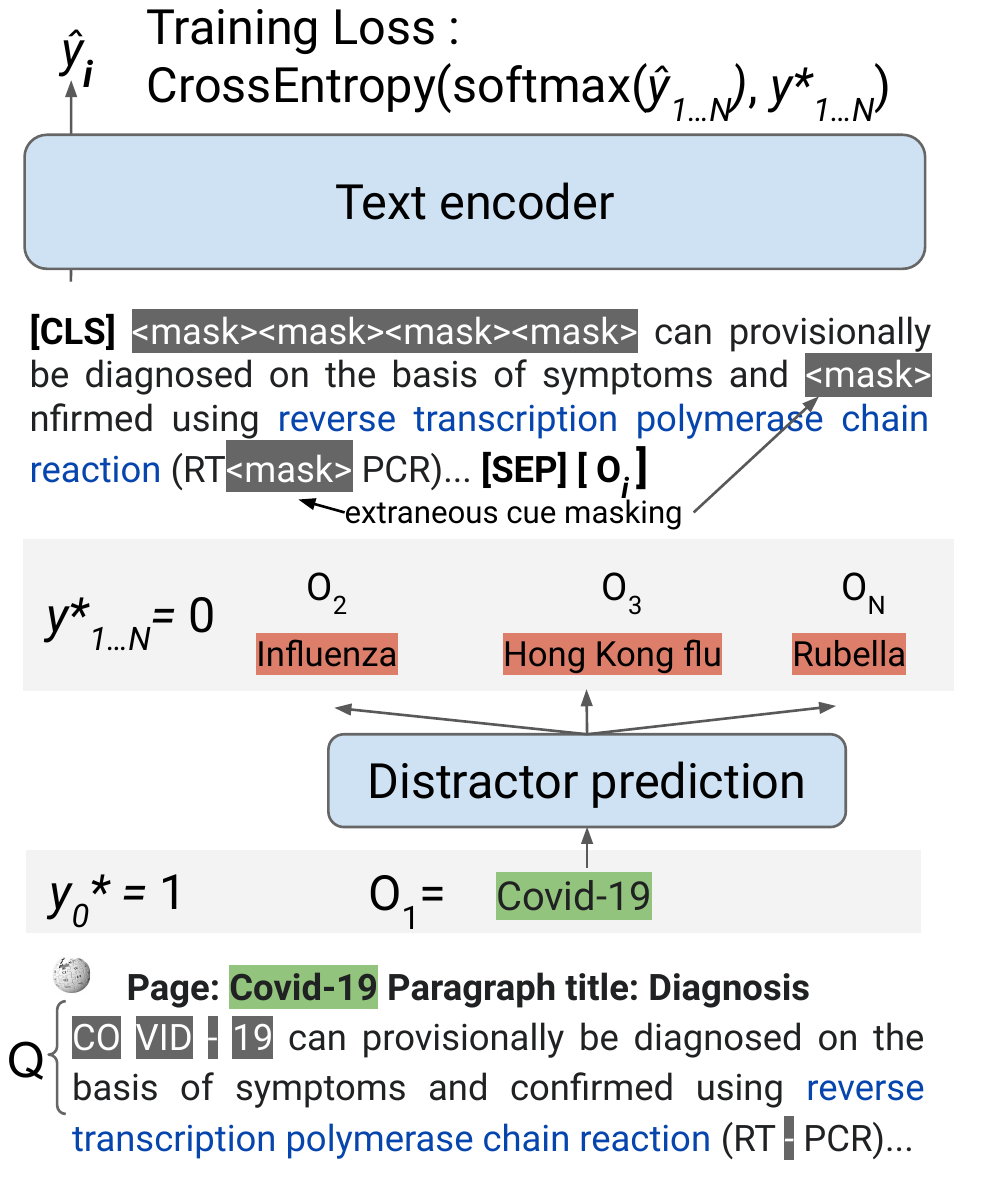}
  \caption{Overview of our data generation technique. Here, we illustrate the naive cue masking scheme used by \cite{jin2020disease}, where tokens from the correct answers are masked. The second maskings  of \textit{co} and - can be cues to the answer.  \label{fig:overview} }
\end{figure}

\section{Improving knowledge infusion \label{sec:infusion}}

DiseaseBERT infuses knowledge by predicting title tokens based on paragraphs where title tokens are masked. We replace that stage with a fine-tuning stage with synthetic MCQA data and provide new analyses on token masking. Figure \ref{fig:overview} illustrates the MCQA generation process and the problem of extraneous masking.
\subsection{MCQA data generation with differential diagnoses and distractor retrieval \label{sec:datagen}}
We generate data by using paragraphs content as questions Q, the title as a correct option O$^*$, and to generate interesting data, we look for related but distinct options, and we mask the direct cues to the correct options in Q.
We noticed that some Wikipedia pages were associated with differential diagnoses cross-linked on DBPedia. We collect associations between pages and differential diagnoses and use them as negative examples. In section \ref{sec:expe}, we will show that they constitute high-quality negative examples. Since these differential diagnoses are not available for all pages, - especially pages that are not related to diseases, but procedures -, we  derive negative examples by using a retrieval model. We will train a retrieval model on \blue{previous,    smaller} MCQA datasets, and evaluate the retriever's ability to find differential diagnoses. We then use the retrieval models to find the most related titles on the same encyclopedia. 

We also experimented with distractor generation by using generative models, but obtained unconvincing results. One advantage of using retrieval, is that because of editorial choices, each page covers different information. This prevents retrieved negative examples from being too similar to the correct answer, and this feature is not exploited by distractor LM-based generations models.

\subsection{Cue masking}

\paragraph{Naive masking} \cite{jin2020disease} masks tokens that are in the answer. However, masking some particular tokens is not necessarily concealing the information about the targeted disease. Some specific tokens are masked everywhere (e.g., a dash -). If these tokens can be predicted based on surrounding terms, the correct disease can be guessed without actually using useful medical knowledge. We call the masking that helps easy guesses \textit{extraneous masking}, an example of which is illustrated in Figure \ref{fig:overview}, where the masked dash can give away the answer if the model has learned that a dash is plausible between \textit{RT} and \textit{PCR}. To address this problem, we will evaluate word-level masking, which necessarily leads to less extraneous masking.  
\paragraph{Probability-matching masking (ours)} \blue{Another problem with naive masking is that if tokens from the correct answer are necessarily masked, a model can detect an incorrect option when it contains a word $w$ that is in the question as $w$ would be masked if the option was correct. We propose a new strategy that takes advantage of the negative examples to alleviate this phenomena. Instead of always masking a word when it is in the correct answer,} we mask a word $w$ with the following probability: 
\vspace{-0.2cm}
\begin{equation}
p_w=  \frac{1}{|\{O_i, w\in O_i, i\in {1..N} \}|}
\end{equation}

where O denotes the other options. This masking scheme ensures that no cue-based classifier can predict the correct answer based on \blue{neither presence nor absence of specific tokens}. It also prevents common tokens from being unnecessarily masked.

\section{Datasets}

\subsection{Pretraining data \label{sec:pretrainingdata}}

We collect new pretraining data from three open-source websites:
\paragraph{Wikipedia Medicine Portal} We crawl pages from the Wikipedia Medicine projects which indexes medical pages.\footnotehref{https://en.wikipedia.org/wiki/Category:All_WikiProject_Medicine_articles}{en.wikipedia.org/...medicine\_articles} We remove pages that match persons or organizations according to WikiData, and pages referring to years. We obtain a total of $75k$ paragraphs.

\paragraph{Wikidoc} We crawl overview pages from the WikiDoc specialized encyclopedia\footnotehref{https://www.wikidoc.org/index.php?title=Special:AllPages}{www.wikidoc.org/...:AllPages} which leads to $28k$ paragraphs.

\paragraph{WikEM} We crawl content pages from the the WikEM encyclopedia\footnotehref{https://wikem.org/w/index.php?title=Special:AllPages&hideredirects=1}{wikem.org/...?title=Special:AllPages}
which is an open source medical encyclopedia targeted for emergency medicine, and we obtain $15k$ paragraphs.

\subsection{Downstream tasks \label{sec:downstream}}

\begin{table}
\small
\begin{tabular}{lr}
\toprule
              Loss/Masking strategy &  \hspace{-2.2cm} MedQA-USMLE Accuracy \\
\midrule
BioLinkBERT-base \citep{yasunaga2022linkbert} &  40.0 \\
\midrule
                         \hspace{0.5cm}+MLM/Token \citep{jin2020disease} &  \hspace{-1.8cm} 40.1 \\
                           \hspace{0.5cm}+MLM/Word &  \hspace{-0.8cm} 41.2 \\
                 \hspace{0.5cm}+Discriminative/Word &  \hspace{-0.8cm} 42.5 \\
 \hspace{0.5cm}+Discriminative/Probability-Matching & \hspace{-0.8cm}  \textbf{43.6} \\
\hspace{1cm}- Differential diagnoses & \hspace{-0.8cm}  42.6 \\
\bottomrule
\end{tabular}

\caption{MEDQA-USMLE validation-set accuracy percentage of BioLinkBERT after knowledge infusion, with varying losses and masking strategies. \vspace{-0.3cm} \label{tab:ablation}}
\end{table}

\begin{table*}
\centering
\small
\begin{tabular}{llllll}
\toprule
                                          &     HEAD-QA+K &       MedMCQA &     MedMCQA+K\hspace{-0.1cm} &   MedQA-USMLE &          MMLU \\
\midrule
                Previous state-of-the-art &          42.4 &          40.0 &          43.0 &          44.6 &          50.7 \\
                \midrule
                         BioLinkBERT-base \cite{yasunaga2022linkbert}\hspace{-0.1cm} &          40.0 &          43.4 &          49.1 &          40.6 &          44.5 \\
\blue{BioLinkBERT-large} \cite{yasunaga2022linkbert} &      \blue{44.1} &       \blue{48.3} &      \blue{52.8} &        \blue{44.6} & \blue{50.6} \\
\blue{BioLinkBERT-base+DiseaseBERT} &\blue{37.8} &  \blue{47.8} &  \blue{44.5} &  \blue{41.2} & \blue{50.5} \\
               BioLinkBERT-base+WikiMedQA &          41.0 &          46.9 &          49.0 &          45.7 &          49.6 \\
              BioLinkBERT-large+WikiMedQA & \textbf{44.5} & \textbf{50.8} & \textbf{53.9} & \textbf{47.2} & \textbf{51.1} \\
\bottomrule
\end{tabular}
\caption{MCQA test accuracy of previous state-of-the-art models and ours. \blue{DiseaseBERT \cite{he-etal-2020-infusing} uses word-level masking}. $\mathcal{D}$+K refers to the dataset $\mathcal{D}$ with retrieved external knowledge concatenated to the question (see section \ref{sec:aug}). HEAD-QA previous sota is from  \cite{liu2020interpretable}. MedMCQA sota are from \cite{pmlr-v174-pal22a}, and MedQA and MMLU sota are from \cite{yasunaga2022linkbert}; these sota results use large-sized models).\vspace{-0.3cm}
\label{tab:downstream} }
\end{table*}

We use 4 medical MCQA datasets to perform evaluation. These datasets contain a question and four options, one of them being correct.

\paragraph{MedQA-USMLE} \cite{jin2020disease} gathers
10k/1.2k/1.2k train/validation/test medical MCQA examples collected from training questions for medical entrance exams found on the Web.
\paragraph{MedMCQA} \cite{pmlr-v174-pal22a} contains 182k/6.2k/4.2k  train/validation/test medical entrance exam training questions.
\paragraph{HEAD-QA \cite{vilares-gomez-rodriguez-2019-head}} we focus on the medical questions translated to English with 0.2k 0.4k validation/test examples, and use the MedMCQA train set as our train set.

\paragraph{MMLU} \cite{hendryckstest2021} is the professional medicine subset of the MMLU language understanding benchmark, which contains 272 test examples. Following \citep{yasunaga2022linkbert}, we used MedQA-USMLE as a training set.

\section{Experiments \label{sec:expe}}

\blue{We generate MCQA examples with the aggregated paragraphs of the pretraining data from Section \ref{sec:pretrainingdata}, with the distractor generation of section and masking strategies of section \ref{sec:infusion}.} We first compare cue masking schemes and pretraining objectives, with ablations on the MedQA-USMLE dataset, then show overall results with the other datasets. We fine-tune BioLinkBERT\footnote{
\blue{WikiMedQA fine-tuning also improves the accuracy of PubMedBERT \cite{gu2021domain} and BioElectra \cite{kanakarajan-etal-2021-bioelectra}
 but both still underperform BioLinkBERT on downstream tasks}}
 \blue{on WikiMedQA then evaluate BioLinkBERT+WikiMedQA on each downstream task with fine-tuning.} We always use standard hyperparameters (5 epochs, sequence length of 256, learning rate of $2.10^{-5}$ batch size of 16). We use a multiple-choice-question answering setup (we predict logit scores for each option by concatenating the question and the option, then use a softmax and optimize the likelihood of the correct option).
\vspace{-0.1cm}
\subsection{Knowledge augmentation \label{sec:aug}}
We also evaluate the pretrained models in a setting where retrieved external knowledge is concatenated to the question. We index previously mentioned Wikipedia medical articles with a BM25\footnote{We also experimented with Dense Passage Retrieval \cite{karpukhin-etal-2020-dense} but obtained inferior results.} search engine, using  ElasticSearch 8.0 default hyperparameters \cite{robertson2009probabilistic}, and we concatenate the 10 most relevant passages\footnote{The concatenated knowledge is truncated if it leads to overflow of text encoder input maximum sequence length.}.

\subsection{Distractor prediction}

To perform distractor prediction, we optimize the \texttt{MultipleNegatives} ranking loss \cite{Henderson2017EfficientNL}
 on the SentenceBERT framework \cite{reimers-2019-sentence-bert} using a BioLinkBERT-base \cite{yasunaga2022linkbert} text encoder and default parameters. We train the ranking model on the MedMCQA training examples,  using the correct answer as a query, the associated non-correct options as relevant and answers to other questions as irrelevant distractors. We evaluate distractor prediction on $3446$ differential diagnoses collected on DBPedia, and found out that using a disease as a query returns a correct differential diagnosis with  precision@3/recall@3 of $11\%/15\%$\footnote{Random chance scores less than $0.2\%/0.2\%$. \blue{BM25 scores $0.6\%/0.7\%$}}.

 We generate 7 incorrect options for each title associated with the paragraphs of texts from the section \ref{sec:pretrainingdata}, and we use probability-matching cue masking to build the WikiMedQA dataset. We use differential diagnoses and retrieved distractors as additional options.

\subsection{Cue masking and distractors retrieval}

Table \ref{tab:ablation} compares fine-tuning on the WikiMedQA Wikipedia part to the DiseaseBERT infusion, and shows the impact of masking strategies. Word-level masking outperforms token-level masking, which shows that less masking leads 
to less extraneous masking and better knowledge infusion. Probability-matching masking also outperforms naive masking at the word level which further validates the importance of addressing extraneous masking. Finally, removing the differential diagnoses from the options substantially decreases accuracy, which showcases the value of differential diagnoses as natural distractors.
\blue{From now on, we refer to the generated data with differential diagnoses and probability-matching masking as WikiMedQA.}

\subsection{Overall results}

Table \ref{tab:downstream} shows the test accuracy of BioLinkBERT fine-tuned on WikiMedQA then on various datasets, compared to the task-specific state-of-the-art. Fine-tuning on WikiMedQA leads to considerable accuracy improvements on all tasks, whether external knowledge is available or not, which shows that this pretraining leads to generalizable text representations for medical question answering.

\vspace{-0.1cm}
\section{Conclusion}
\vspace{-0.1cm}
We proposed a new dataset for Medical MCQA in English pretraining by leveraging distractors retrieval and cue masking. We identified the problem of extraneous masking, proposed the probability-matching masking and demonstrated its advantage, and showed that differential diagnoses were helpful distractors. Fine-tuning on WikiMedQA leads to considerable improvement on several datasets, and this method can be ported to other languages.

\section{Acknowledgements}
This work was supported by the CHISTERA grant of the Call XAI 2019 of the ANR with the grant number Project-ANR-21-CHR4-000.

\bibliography{main}

\begin{thebibliography}{31}
\expandafter\ifx\csname natexlab\endcsname\relax\def\natexlab#1{#1}\fi

\bibitem[{Chen et~al.(2019)Chen, Zhou, Shi, Fan, and Luo}]{rwextractive}
Jun Chen, Jingbo Zhou, Zhenhui Shi, Bin Fan, and Chengliang Luo. 2019.
\newblock \href {https://doi.org/10.1109/BIBM47256.2019.8982973} {Knowledge
  abstraction matching for medical question answering}.
\newblock In \emph{2019 IEEE International Conference on Bioinformatics and
  Biomedicine (BIBM)}, pages 342--347.

\bibitem[{Chung et~al.(2020)Chung, Chan, and Fan}]{chung-etal-2020-BERT}
Ho-Lam Chung, Ying-Hong Chan, and Yao-Chung Fan. 2020.
\newblock \href {https://www.aclweb.org/anthology/2020.findings-emnlp.393} {A
  {BERT}-based distractor generation scheme with multi-tasking and negative
  answer training strategies.}
\newblock In \emph{Proceedings of the 2020 Conference on Empirical Methods in
  Natural Language Processing: Findings}, pages 4390--4400, Online. Association
  for Computational Linguistics.

\bibitem[{Donnelly et~al.(2006)}]{donnelly2006snomed}
Kevin Donnelly et~al. 2006.
\newblock Snomed-ct: The advanced terminology and coding system for ehealth.
\newblock \emph{Studies in health technology and informatics}, 121:279.

\bibitem[{Gu et~al.(2021)Gu, Tinn, Cheng, Lucas, Usuyama, Liu, Naumann, Gao,
  and Poon}]{gu2021domain}
Yu~Gu, Robert Tinn, Hao Cheng, Michael Lucas, Naoto Usuyama, Xiaodong Liu,
  Tristan Naumann, Jianfeng Gao, and Hoifung Poon. 2021.
\newblock Domain-specific language model pretraining for biomedical natural
  language processing.
\newblock \emph{ACM Transactions on Computing for Healthcare (HEALTH)},
  3(1):1--23.

\bibitem[{Guu et~al.(2020)Guu, Lee, Tung, Pasupat, and Chang}]{Realm}
Kelvin Guu, Kenton Lee, Zora Tung, Panupong Pasupat, and Ming-Wei Chang. 2020.
\newblock Realm: Retrieval-augmented language model pre-training.
\newblock In \emph{Proceedings of the 37th International Conference on Machine
  Learning}, ICML'20. JMLR.org.

\bibitem[{Ha and Yaneva(2018)}]{ha-yaneva-2018-automatic}
Le~An Ha and Victoria Yaneva. 2018.
\newblock \href {https://doi.org/10.18653/v1/W18-0548} {Automatic distractor
  suggestion for multiple-choice tests using concept embeddings and information
  retrieval}.
\newblock In \emph{Proceedings of the Thirteenth Workshop on Innovative Use of
  {NLP} for Building Educational Applications}, pages 389--398, New Orleans,
  Louisiana. Association for Computational Linguistics.

\bibitem[{He et~al.(2020)He, Zhu, Zhang, Chen, and
  Caverlee}]{he-etal-2020-infusing}
Yun He, Ziwei Zhu, Yin Zhang, Qin Chen, and James Caverlee. 2020.
\newblock \href {https://doi.org/10.18653/v1/2020.emnlp-main.372} {{I}nfusing
  {D}isease {K}nowledge into {BERT} for {H}ealth {Q}uestion {A}nswering,
  {M}edical {I}nference and {D}isease {N}ame {R}ecognition}.
\newblock In \emph{Proceedings of the 2020 Conference on Empirical Methods in
  Natural Language Processing (EMNLP)}, pages 4604--4614, Online. Association
  for Computational Linguistics.

\bibitem[{Henderson et~al.(2017)Henderson, Al-Rfou, Strope, Sung, Luk{\'a}cs,
  Guo, Kumar, Miklos, and Kurzweil}]{Henderson2017EfficientNL}
Matthew Henderson, Rami Al-Rfou, Brian Strope, Yun-Hsuan Sung, L{\'a}szl{\'o}
  Luk{\'a}cs, Ruiqi Guo, Sanjiv Kumar, Balint Miklos, and Ray Kurzweil. 2017.
\newblock Efficient natural language response suggestion for smart reply.
\newblock \emph{ArXiv}, abs/1705.00652.

\bibitem[{Hendrycks et~al.(2021)Hendrycks, Burns, Basart, Zou, Mazeika, Song,
  and Steinhardt}]{hendryckstest2021}
Dan Hendrycks, Collin Burns, Steven Basart, Andy Zou, Mantas Mazeika, Dawn
  Song, and Jacob Steinhardt. 2021.
\newblock Measuring massive multitask language understanding.
\newblock \emph{Proceedings of the International Conference on Learning
  Representations (ICLR)}.

\bibitem[{Jin et~al.(2020)Jin, Pan, Oufattole, Weng, Fang, and
  Szolovits}]{jin2020disease}
Di~Jin, Eileen Pan, Nassim Oufattole, Wei-Hung Weng, Hanyi Fang, and Peter
  Szolovits. 2020.
\newblock What disease does this patient have? a large-scale open domain
  question answering dataset from medical exams.
\newblock \emph{arXiv preprint arXiv:2009.13081}.

\bibitem[{Jin et~al.(2019)Jin, Dhingra, Liu, Cohen, and
  Lu}]{jin-etal-2019-pubmedqa}
Qiao Jin, Bhuwan Dhingra, Zhengping Liu, William Cohen, and Xinghua Lu. 2019.
\newblock \href {https://doi.org/10.18653/v1/D19-1259} {{P}ub{M}ed{QA}: A
  dataset for biomedical research question answering}.
\newblock In \emph{Proceedings of the 2019 Conference on Empirical Methods in
  Natural Language Processing and the 9th International Joint Conference on
  Natural Language Processing (EMNLP-IJCNLP)}, pages 2567--2577, Hong Kong,
  China. Association for Computational Linguistics.

\bibitem[{Kanakarajan et~al.(2021)Kanakarajan, Kundumani, and
  Sankarasubbu}]{kanakarajan-etal-2021-bioelectra}
Kamal~raj Kanakarajan, Bhuvana Kundumani, and Malaikannan Sankarasubbu. 2021.
\newblock \href {https://doi.org/10.18653/v1/2021.bionlp-1.16}
  {{B}io{ELECTRA}:pretrained biomedical text encoder using discriminators}.
\newblock In \emph{Proceedings of the 20th Workshop on Biomedical Language
  Processing}, pages 143--154, Online. Association for Computational
  Linguistics.

\bibitem[{Karpukhin et~al.(2020)Karpukhin, Oguz, Min, Lewis, Wu, Edunov, Chen,
  and Yih}]{karpukhin-etal-2020-dense}
Vladimir Karpukhin, Barlas Oguz, Sewon Min, Patrick Lewis, Ledell Wu, Sergey
  Edunov, Danqi Chen, and Wen-tau Yih. 2020.
\newblock \href {https://doi.org/10.18653/v1/2020.emnlp-main.550} {Dense
  passage retrieval for open-domain question answering}.
\newblock In \emph{Proceedings of the 2020 Conference on Empirical Methods in
  Natural Language Processing (EMNLP)}, pages 6769--6781, Online. Association
  for Computational Linguistics.

\bibitem[{Lee et~al.(2020)Lee, Yoon, Kim, Kim, Kim, So, and
  Kang}]{lee2020biobert}
Jinhyuk Lee, Wonjin Yoon, Sungdong Kim, Donghyeon Kim, Sunkyu Kim, Chan~Ho So,
  and Jaewoo Kang. 2020.
\newblock Biobert: a pre-trained biomedical language representation model for
  biomedical text mining.
\newblock \emph{Bioinformatics}, 36(4):1234--1240.

\bibitem[{Lewis et~al.(2020)Lewis, Ott, Du, and
  Stoyanov}]{lewis-etal-2020-pretrained}
Patrick Lewis, Myle Ott, Jingfei Du, and Veselin Stoyanov. 2020.
\newblock \href {https://www.aclweb.org/anthology/2020.clinicalnlp-1.17}
  {Pretrained language models for biomedical and clinical tasks: Understanding
  and extending the state-of-the-art}.
\newblock In \emph{Proceedings of the 3rd Clinical Natural Language Processing
  Workshop}, pages 146--157, Online. Association for Computational Linguistics.

\bibitem[{Li et~al.(2021)Li, Zhong, and Chen}]{li-etal-2021-mlec}
Jing Li, Shangping Zhong, and Kaizhi Chen. 2021.
\newblock \href {https://doi.org/10.18653/v1/2021.emnlp-main.698} {{MLEC-QA}:
  {A} {C}hinese {M}ulti-{C}hoice {B}iomedical {Q}uestion {A}nswering
  {D}ataset}.
\newblock In \emph{Proceedings of the 2021 Conference on Empirical Methods in
  Natural Language Processing}, pages 8862--8874, Online and Punta Cana,
  Dominican Republic. Association for Computational Linguistics.

\bibitem[{Liu et~al.(2020)Liu, Chowdhury, Zhang, Caragea, and
  Yu}]{liu2020interpretable}
Ye~Liu, Shaika Chowdhury, Chenwei Zhang, Cornelia Caragea, and Philip~S Yu.
  2020.
\newblock Interpretable multi-step reasoning with knowledge extraction on
  complex healthcare question answering.
\newblock \emph{arXiv preprint arXiv:2008.02434}.

\bibitem[{Michalopoulos et~al.(2021)Michalopoulos, Wang, Kaka, Chen, and
  Wong}]{michalopoulos-etal-2021-umlsbert}
George Michalopoulos, Yuanxin Wang, Hussam Kaka, Helen Chen, and Alexander
  Wong. 2021.
\newblock \href {https://doi.org/10.18653/v1/2021.naacl-main.139}
  {{U}mls{BERT}: Clinical domain knowledge augmentation of contextual
  embeddings using the {U}nified {M}edical {L}anguage {S}ystem
  {M}etathesaurus}.
\newblock In \emph{Proceedings of the 2021 Conference of the North American
  Chapter of the Association for Computational Linguistics: Human Language
  Technologies}, pages 1744--1753, Online. Association for Computational
  Linguistics.

\bibitem[{Pal et~al.(2022)Pal, Umapathi, and Sankarasubbu}]{pmlr-v174-pal22a}
Ankit Pal, Logesh~Kumar Umapathi, and Malaikannan Sankarasubbu. 2022.
\newblock \href {https://proceedings.mlr.press/v174/pal22a.html} {Medmcqa: A
  large-scale multi-subject multi-choice dataset for medical domain question
  answering}.
\newblock In \emph{Proceedings of the Conference on Health, Inference, and
  Learning}, volume 174 of \emph{Proceedings of Machine Learning Research},
  pages 248--260. PMLR.

\bibitem[{Pampari et~al.(2018)Pampari, Raghavan, Liang, and
  Peng}]{pampari-etal-2018-emrqa}
Anusri Pampari, Preethi Raghavan, Jennifer Liang, and Jian Peng. 2018.
\newblock \href {https://doi.org/10.18653/v1/D18-1258} {emr{QA}: A large corpus
  for question answering on electronic medical records}.
\newblock In \emph{Proceedings of the 2018 Conference on Empirical Methods in
  Natural Language Processing}, pages 2357--2368, Brussels, Belgium.
  Association for Computational Linguistics.

\bibitem[{Petroni et~al.(2019)Petroni, Rockt{\"a}schel, Riedel, Lewis, Bakhtin,
  Wu, and Miller}]{petroni-etal-2019-language}
Fabio Petroni, Tim Rockt{\"a}schel, Sebastian Riedel, Patrick Lewis, Anton
  Bakhtin, Yuxiang Wu, and Alexander Miller. 2019.
\newblock \href {https://doi.org/10.18653/v1/D19-1250} {Language models as
  knowledge bases?}
\newblock In \emph{Proceedings of the 2019 Conference on Empirical Methods in
  Natural Language Processing and the 9th International Joint Conference on
  Natural Language Processing (EMNLP-IJCNLP)}, pages 2463--2473, Hong Kong,
  China. Association for Computational Linguistics.

\bibitem[{Phang et~al.(2018)Phang, F{\'e}vry, and Bowman}]{phang2018sentence}
Jason Phang, Thibault F{\'e}vry, and Samuel~R Bowman. 2018.
\newblock Sentence encoders on stilts: Supplementary training on intermediate
  labeled-data tasks.
\newblock \emph{arXiv preprint arXiv:1811.01088}.

\bibitem[{Reimers and Gurevych(2019)}]{reimers-2019-sentence-bert}
Nils Reimers and Iryna Gurevych. 2019.
\newblock \href {https://arxiv.org/abs/1908.10084} {Sentence-bert: Sentence
  embeddings using siamese bert-networks}.
\newblock In \emph{Proceedings of the 2019 Conference on Empirical Methods in
  Natural Language Processing}. Association for Computational Linguistics.

\bibitem[{Robertson et~al.(2009)Robertson, Zaragoza
  et~al.}]{robertson2009probabilistic}
Stephen Robertson, Hugo Zaragoza, et~al. 2009.
\newblock The probabilistic relevance framework: Bm25 and beyond.
\newblock \emph{Foundations and Trends{\textregistered} in Information
  Retrieval}, 3(4):333--389.

\bibitem[{Rogers et~al.(2021)Rogers, Gardner, and Augenstein}]{rogers2021qa}
Anna Rogers, Matt Gardner, and Isabelle Augenstein. 2021.
\newblock Qa dataset explosion: A taxonomy of nlp resources for question
  answering and reading comprehension.
\newblock \emph{arXiv preprint arXiv:2107.12708}.

\bibitem[{Schulz and Hahn(2001)}]{SCHULZ2001207}
Stefan Schulz and Udo Hahn. 2001.
\newblock \href {https://doi.org/https://doi.org/10.1016/S1386-5056(01)00201-5}
  {Medical knowledge reengineering—converting major portions of the umls into
  a terminological knowledge base}.
\newblock \emph{International Journal of Medical Informatics}, 64(2):207--221.

\bibitem[{Schuyler et~al.(1993)Schuyler, Hole, Tuttle, and Sherertz}]{umls93}
P~Schuyler, W~Hole, Mark Tuttle, and David Sherertz. 1993.
\newblock The umls metathesaurus: representing different views of biomedical
  concepts.
\newblock \emph{Bulletin of the Medical Library Association}, 81:217--22.

\bibitem[{Vilares and
  G{\'o}mez-Rodr{\'i}guez(2019)}]{vilares-gomez-rodriguez-2019-head}
David Vilares and Carlos G{\'o}mez-Rodr{\'i}guez. 2019.
\newblock \href {https://doi.org/10.18653/v1/P19-1092} {{HEAD}-{QA}: A
  healthcare dataset for complex reasoning}.
\newblock In \emph{Proceedings of the 57th Annual Meeting of the Association
  for Computational Linguistics}, pages 960--966, Florence, Italy. Association
  for Computational Linguistics.

\bibitem[{Wang et~al.(2021)Wang, Tang, Duan, Wei, Huang, Ji, Cao, Jiang, and
  Zhou}]{kadapter}
Ruize Wang, Duyu Tang, Nan Duan, Zhongyu Wei, Xuanjing Huang, Jianshu Ji,
  Guihong Cao, Daxin Jiang, and Ming Zhou. 2021.
\newblock \href {https://doi.org/10.18653/v1/2021.findings-acl.121}
  {{K-Adapter}: {I}nfusing {K}nowledge into {P}re-{T}rained {M}odels with
  {A}dapters}.
\newblock In \emph{Findings of the Association for Computational Linguistics:
  ACL-IJCNLP 2021}, pages 1405--1418, Online. Association for Computational
  Linguistics.

\bibitem[{Xia et~al.(2021)Xia, Wang, Shi, Zhou, Lu, Huang, and
  Xiong}]{rwknowledge}
Yuan Xia, Chunyu Wang, Zhenhui Shi, Jingbo Zhou, Chao Lu, Haifeng Huang, and
  Hui Xiong. 2021.
\newblock Medical entity relation verification with large-scale machine reading
  comprehension.
\newblock In \emph{Proceedings of the 27th ACM SIGKDD Conference on Knowledge
  Discovery \& Data Mining (KDD)}, pages 3765--3774.

\bibitem[{Yasunaga et~al.(2022)Yasunaga, Leskovec, and
  Liang}]{yasunaga2022linkbert}
Michihiro Yasunaga, Jure Leskovec, and Percy Liang. 2022.
\newblock Linkbert: Pretraining language models with document links.
\newblock In \emph{Association for Computational Linguistics (ACL)}.

\end{thebibliography}
\bibliographystyle{acl_natbib}


\end{document}